\documentclass[manuscript,screen]{acmart}
\usepackage{multirow}
\usepackage{adjustbox}
\usepackage[utf8]{inputenc}
\usepackage{array}
\usepackage{mathtools}
\usepackage{makecell}
\usepackage{amsmath}
\usepackage{bm}

\usepackage[edges]{forest}
\tikzset{%
    parent/.style =          {align=center,text width=2cm,rounded corners=3pt, line width=0.3mm, fill=gray!10,draw=gray!80},
    child/.style =           {align=center,text width=2.3cm,rounded corners=3pt, fill=blue!10,draw=blue!80,line width=0.3mm},
    grandchild/.style =      {align=center,text width=2cm,rounded corners=3pt},
    greatgrandchild/.style = {align=center,text width=1.5cm,rounded corners=3pt},
    greatgrandchild2/.style = {align=center,text width=1.5cm,rounded corners=3pt},    
    referenceblock/.style =  {align=center,text width=1.5cm,rounded corners=2pt},
    pretrain/.style =           {align=center,text width=2.2cm,rounded corners=3pt, fill=blue!10,draw=blue!80,line width=0.3mm},   
    pretrain_work/.style =           {align=center, text width=3cm,rounded corners=3pt, fill=blue!10,draw=blue!0,line width=0.3mm},  
    template/.style =           {align=center,text width=2.2cm,rounded corners=3pt, fill=red!10,draw=red!80,line width=0.3mm},   
    template_work/.style =           {align=center,text width=3cm,rounded corners=3pt, fill=red!10,draw=red!0,line width=0.3mm},             
}

\AtBeginDocument{%
  \providecommand\BibTeX{{%
    \normalfont B\kern-0.5em{\scshape i\kern-0.25em b}\kern-0.8em\TeX}}}

\begin{document}

\title{Graph Neural Networks for Text Classification: A Survey}

\author{Kunze Wang}
\author{Yihao Ding}
\affiliation{%
  \institution{The University of Sydney}
  \country{Australia}
}
\author{Soyeon Caren Han}
\affiliation{%
\institution{The University of Western Australia},
\institution{The University of Sydney}
\country{Australia}
}



\renewcommand{\shortauthors}{Kunze, et al.}

\begin{abstract}
Text Classification is the most essential and fundamental problem in Natural Language Processing. While numerous recent text classification models applied the sequential deep learning technique, graph neural network-based models can directly deal with complex structured text data and exploit global information. Many real text classification applications can be naturally cast into a graph, which captures words, documents, and corpus global features. 
In this survey, we bring the coverage of methods up to 2023, including corpus-level and document-level graph neural networks. We discuss each of these methods in detail, dealing with the graph construction mechanisms and the graph-based learning process. As well as the technological survey, we look at issues behind and future directions addressed in text classification using graph neural networks. We also cover datasets, evaluation metrics, and experiment design and present a summary of published performance on the publicly available benchmarks. Note that we present a comprehensive comparison between different techniques and identify the pros and cons of various evaluation metrics in this survey.

\end{abstract}



\maketitle

\section{Introduction}\label{sec:intro}








Text classification aims to classify a given document into certain pre-defined classes, and is considered as a fundamental task in Natural Language Processing (NLP). It includes a large number of downstream tasks, such as topic classification\cite{zhang2015character}, and sentiment analysis\cite{tai2015improved}. Traditional text classification methods build representation on the text using N-gram\cite{cavnar1994n} or Term Frequency-Inverse Document Frequency (TF-IDF) \cite{hakim2014automated} and apply traditional machine learning models, such as SVM\cite{joachims2005text}, to classify the documents. With the development of neural networks, more deep learning models have been applied to text classification, including convolutional neural networks (CNN)\cite{kim2014convolutional}, recurrent neural networks (RNN)\cite{tang2015document} and attention-based\cite{vaswani2017attention} models and large language models\cite{devlin2018bert}.

However, these methods are either unable to handle the complex relationships between words and documents\cite{TEXTGCN2019}, and can not efficiently explore the contextual-aware word relations\cite{TEXTING2020}. To resolve such obstacles, graph neural networks (GNN) are introduced. GNN is used with graph-structure datasets so a graph needs to be built for text classification. There are two main approaches to constructing graphs, corpus-level graphs and document-level graphs. The datasets are either built into single or multiple corpus-level graphs representing the whole corpus or numerous document-level graphs and each of them represents a document. The corpus-level graph can capture the global structural information of the entire corpus, while the document-level graph can capture the word-to-word relationships within a document explicitly. Both ways of applying graph neural networks to text classification achieve good performance.

This paper mainly focuses on GNN-based text classification techniques, datasets, and their performance. The graph construction approaches for both corpus-level and document-level graphs are addressed in detail. Papers on the following aspects will be reviewed:

\begin{itemize}
    \item GNNs-based text classification approaches. Papers that design GNN-based frameworks to enhance the feature representation or directly apply GNNs to conduct sequence text classification tasks will be summarized, described and discussed. GNNs applied for token-level classification (Natural Language Understanding) tasks, including NER, slot filling, etc, will not be discussed in this work. 
    \item Text classification benchmark datasets and their performance applied by GNN-based models. The text classification datasets with commonly used metrics used by GNNs-based text classification models will be summarized and categorized based on task types, together with the model performance on these datasets.
\end{itemize}

\subsection{Related Surveys and Our Contribution}

\begin{figure*}
\footnotesize
\begin{forest}
    for tree={
        forked edges,
        grow'=0,
        draw,
        rounded corners,
        node options={align=center},
        text width=2.7cm,
        s sep=6pt,
        calign=child edge, 
        calign child=(n_children()+1)/2
    }
    [GNN for Text Classification, fill=gray!45, parent
        [Corpus-level Graph, for tree={ pretrain}
            [Word and Document nodes,  pretrain
                [PMI+TF-IDF,  pretrain
                    [TextGCN\cite{TEXTGCN2019}; SGC\cite{SGC2019}; S$^2$GC\cite{S2GCN2020}; NMGC\cite{NMGC2021}; TG-Transformer\cite{TGTRANSFORMER2020}; BertGCN\cite{BERTGCN2021}, pretrain_work]
                ]
                [Multi-Graph/Multi-Dimensional Edge, pretrain
                    [TensorGCN\cite{TENSORGCN2020}; ME-GCN\cite{MEGCN2022},pretrain_work]
                ]
                [Inductive Learning, pretrain
                    [HeteGCN\cite{HETEGCN2021}; InducT-GCN\cite{wang2022induct}; T-VGAE\cite{TVGAE2021}, style = pretrain_work]
                ]
            ]
            [Document Nodes, pretrain
                [knn-GCN\cite{benamira2019semi}; TextGTL\cite{TEXTGTL2021},style = pretrain_work]
            ]
            [Word Nodes, pretrain
                [VGCN-BERT\cite{VGCNBERT2020}, style = pretrain_work]
            ]
            [Extra Topic Nodes, pretrain
                [Single Layer topic nodes, pretrain
                    [HGAT\cite{HGAT2019}; STGCN\cite{STGCN2020},style = pretrain_work]
                ]
                [Multi-layer Topic Node, pretrain
                    [DHTG\cite{DHTG2020}, style = pretrain_work]
                ]
            ]
        ]
        [Document-Level Graph, for tree={template}
            [Local word consecutive,  template
                [Simple consecutive graph models, template       
                    [Text-Level-GNN\cite{TEXTLEVELGNN2019}; MPAD\cite{MPAD2020}; TextING\cite{TEXTING2020}, template_work]                         
                ]
                [Advanced graph models, template  
                    [MLGNN\cite{MLGNN2021}; DADGNN\cite{DADGNN2021}; TextSSL\cite{TEXTSSL2021}, template_work]      
                ]
            ]
            [Global Word Co-occurrence,  template
                [Only co-occurrence, template
                  [DAGNN\cite{DAGNN2019},template_work] 
                ]
                [With Extra Edges, template
                  [ReGNN\cite{REGNN2019}; GFN\cite{GFN},template_work] 
                ]
            ]
            [Other Word Graphs,  template
                 [HyperGAT\cite{HYPERGAT2020}; IGCN\cite{IGCN2020}; GTNT\cite{GTNT2021},template_work] 
            ]
        ]
    ]
\end{forest}
\caption{Categorizing the graph neural network text classification models.}
\label{fig:mindmap}
\end{figure*}
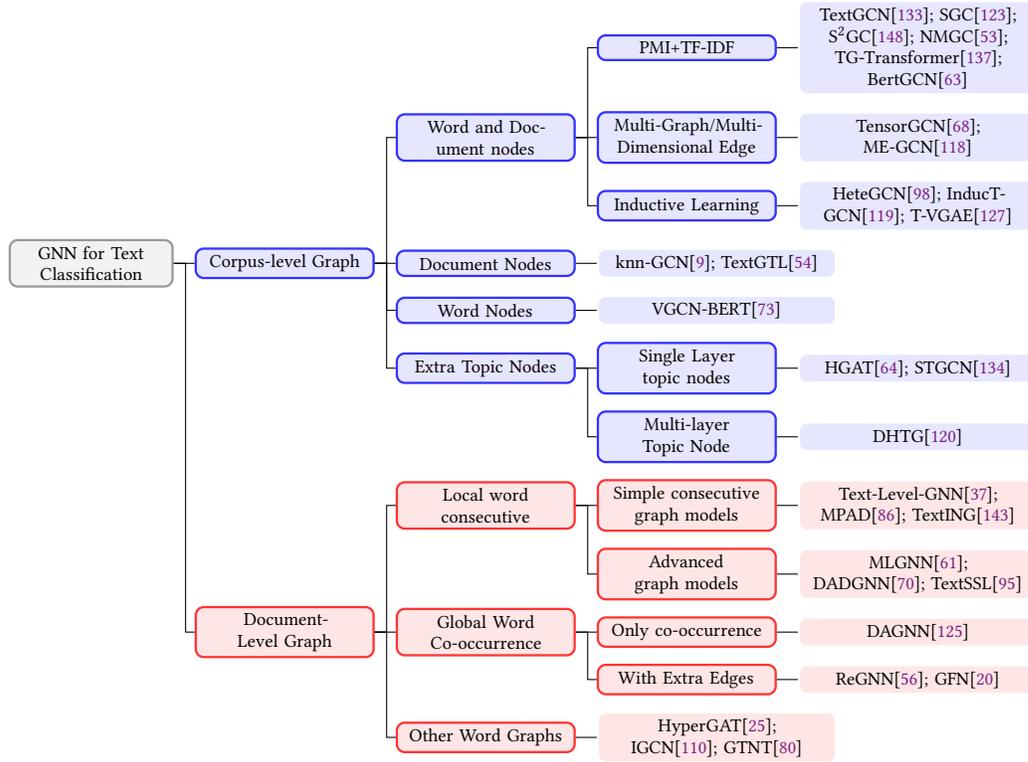

Before 2019, the text classification survey papers \cite{xing2010brief, khan2010review, harish2010representation, aggarwal2012survey, vijayan2017comprehensive} have focused on covering traditional machine learning-based text classification models. 
Recently, with the rapid development of deep learning techniques, \cite{minaee2021deep, zulqarnain2020comparative, zhou2020review, li2022survey} review the various deep learning based approaches. In addition, some papers not only review the SoTA model architectures, but summarize the overall workflow \cite{jindal2015techniques, kadhim2019survey, mironczuk2018recent, kowsari2019text, bhavani2021review} or specific techniques for text classification including word embedding \cite{selva2021review}, feature selection \cite{deng2019feature, shah2016review, pintas2021feature} , term weighting \cite{patra2013survey, alsaeedi2020survey} and etc. Meanwhile, some growing potential text classification architectures are surveyed, such as CNNs \cite{yang2016hierarchical}, attention mechanisms \cite{mariyam2021literature}. Since the powerful ability to represent non-Euclidean relation, GNNs have been used in multiple practical fields and reviewed e.g.   financial application \cite{wang2021review}, traffic prediction \cite{liu2021traffic}, bio-informatics \cite{zhang2021graph}, power system \cite{liao2021review}, recommendation system \cite{gao2022graph, liang2021survey, yang2021review}. Moreover, \cite{bronstein2017geometric, battaglia2018relational, zhang2019graph, zhou2020graph, wu2020comprehensive} comprehensively review the general algorithms and applications of GNNs, as well as certain surveys mainly focus on specific perspectives including graph construction \cite{skarding2021foundations, thomas2022graph}, graph representation \cite{hamilton2017representation}, training \cite{xie2022self}, pooling \cite{liu2022graph} and more. However, only \cite{minaee2021deep, li2022survey} briefly introduce certain SoTA GNN-based text classification models. A recent short review paper \cite{malekzadeh2021review} reviews several SoTA models without providing a comprehensive overview in this area.
The contribution of this survey includes:
\begin{itemize}
\item This is the first survey focused only on graph neural networks for text classification with a comprehensive description and critical discussion on more than twenty GNN text classification models. 
\item We categorize the existing GNN text classification models into two main categories with multiple sub-categories, and the tree structure of all the models shows in Figure \ref{fig:mindmap}. 
\item We compare these models in terms of graph construction, node embedding initialization, and graph learning methods. And We also compare the performance of these models on the benchmark datasets and discuss the key findings.
\item We discuss the existing challenges and some potential future work for GNN text classification models.
\end{itemize}

\subsection{Text Classification Tasks}
\hfill\\
Text classification involves assigning a pre-defined label to a given text sequence. The process typically involves encoding pre-processed raw text into numerical representations and using classifiers to predict the corresponding categories. Typical sub-tasks include sentiment analysis, topic labelling, news categorization, and hate speech detection. Certain frameworks can be extended to advanced applications such as information retrieval, summarising, question answering, and natural language inference. This paper focuses specifically on GNN-based models used for typical text classification.
\begin{itemize}
    \item \textbf{Sentiment Analysis} is a task that aims to identify the emotional states and subjective opinions expressed in the input text, such as reviews, micro-blogs, etc. This can be achieved through binary or multi-class classification. Effective sentiment analysis can aid in making informed business decisions based on user feedback.
    \item \textbf{Topic Classification} is a supervised deep learning task to automatically understand the text content and classified into multiple domain-specific categories, typically more than two. The data sources may gather from different domains, including Wikipedia pages, newspapers, scientific papers, etc.
    \item \textbf{Junk Information Detection} involves detecting inappropriate social media content. Social media providers commonly use approaches like hate speech, abusive language, advertising or spam detection to remove such content efficiently.
\end{itemize}

\subsection{Text Classification Development}

Many traditional machine learning methods and deep learning models are selected as the baselines for comparing with the GNN-based text classifiers. We mainly summarized those baselines into three types: 

\textit{\textbf{Traditional Machine Learning}}:
In earlier years, traditional methods such as Support Vector Machines (SVM) \cite{zhang2011comparative} and Logistic Regression \cite{genkin2007large} utilized sparse representations like Bag of Words (BoW) and TF-IDF. However, recent advancements \cite{lilleberg2015support, yin2015document, ren2016topic} have focused on dense representations, such as Word2vec, GloVe, and Fasttext, to mitigate the limitations of sparse representations. These dense representations are also used as inputs for sophisticated methods, such as Deep Averaging Networks (DAN) \cite{iyyer2015deep} and Paragraph Vector (Doc2Vec) \cite{le2014distributed}, to achieve new state-of-the-art results.


\textit{\textbf{Sequential Models}}:
RNNs and CNNs have been utilized to capture local-level semantic and syntactic information of consecutive words from input text bodies. The upgraded models, such as LSTM \cite{graves2012long} and GRU \cite{cho2014learning}, have been proposed to address the vanishing or exploding gradient problems caused by vanilla RNN. CNN-based structures have been applied to capture N-gram features by using one or more convolution and pooling layers, such as Dynamic CNN \cite{kalchbrenner2014convolutional} and TextCNN \cite{kim2014convolutional}. However, these models can only capture local dependencies of consecutive words. To capture longer-term or non-Euclidean relations, improved RNN structures, such as Tree-LSTM \cite{tai2015improved} and MT-LSTM \cite{liu2015multi}, and global semantic information, like TopicRNN \cite{dieng2016topicrnn}, have been proposed. Additionally, graph \cite{peng2018large} and tree structure \cite{mou2015natural} enhanced CNNs have been proposed to learn more about global and long-term dependencies.


\textit{\textbf{Attentions and Transformers}}:
attention mechanisms \cite{bahdanau2014neural} have been widely adopted to capture long-range dependencies, such as hierarchical attention networks \cite{abreu2019hierarchical} and attention-based hybrid models \cite{yang2016hierarchical}. Self-attention-based transformer models have achieved state-of-the-art performance on many text classification benchmarks via pre-training on some tasks to generate strong contextual word representations. However, these models only focus on learning the relation between input text bodies and ignore the global and corpus level information. Researchers have proposed combining the benefits of attention mechanisms and Graph Neural Networks (GNNs) to learn both the relation between input text bodies and the global and corpus level information, such as VGCN-BERT \cite{VGCNBERT2020} and BERTGCN \cite{BERTGCN2021}.

\subsection{Outline}
The outline of this survey is as follows:
\begin{itemize}
    \item Section \ref{sec:intro} presents the research questions and provides an overview of applying Graph Neural Networks to text classification tasks, along with the scope and organization of this survey.
    \item Section \ref{sec:back} provides background information on text classification and graph neural networks and introduces the key concepts of applying GNNs to text classification from a designer's perspective.
    \item Section \ref{sec:corpussec} and Section \ref{sec:docsec} discuss previous work on Corpus-level Graph Neural Networks and Document-level Graph Neural Networks, respectively, and provide a comparative analysis of the strengths and weaknesses of these two approaches.
    \item Section \ref{sec:datasetandmetrics} introduces the commonly used datasets and evaluation metrics in GNN for text classification.
    \item Section \ref{sec:performance} reports the performance of various GNN models on a range of benchmark datasets for text classification and discusses the key findings.
    \item The challenges for the existing methods and some potential future works are discussed in Section \ref{sec:challenges}.
    \item In Section \ref{sec:conclusion}, we present the conclusions of our survey on GNN for text classification and discuss potential directions for future work.
\end{itemize}

\section{Backgrounds of GNN}\label{sec:back}


\subsection{Definition of Graph }
\hfill\\
A graph in this paper is represented as $G =  (V, E)$, where $V$ and $E$ represent a set of nodes (vertices) and edges of $G$, respectively. A single node in the node set is represented $v_{i} \in V$, as well as $e_{ij} = (v_{i},v_{j}) \in E$ donates an edge between node $v_{i}$ and $v_{j}$. The adjacent matrix of graph $G$ is represented as $A$, where $A \in \mathbb{R}^{n \times n}$ and $n$ is the number of nodes in graph $G$. If $e_{ij} \in E$, $A_{ij} = 1$, otherwise $A_{ij} = 0$. In addition, we use $\textit{\textbf{X}}$ and $\textit{\textbf{E}}$ to represent the nodes and edges representations in graph $G$, where $\textit{\textbf{X}} \in \mathbb{R}^{n \times m}$ and $\textit{\textbf{E}} \in \mathbb{R}^{n \times c}$. $\textbf{\textit{x}}_i \in \mathbb{R}^m$ represents the $m$-dimensional vector of node $v_{i}$ and $\textbf{\textit{e}}_{ij} \in \mathbb{R}^c$ represents the $c$-dimensional vector of edge $e_{ij}$ (most of the recent studies set $c=1$ to represent a weighting scalar). \textbf{\textit{A}} donates the edge feature weighted adjacent matrix.

\subsection{Traditional Graph-based Algorithms}
\hfill\\
Before GNNs were broadly used for representing irregular relations, traditional graph-based algorithms have been applied to model the non-Euclidean structures in text classification e.g. Random Walk \cite{szummer2001partially,zhou2005text}, Graph Matching \cite{schenker2004classification, silva2014bog}, Graph Clustering \cite{matsuo2006graph} which has been well summarized in \cite{wu2021graph}. There are three common limitations of those traditional graph-based algorithms. Firstly, most of those algorithms mainly focus on capturing graph-level structure information without considering the significance of node and edge features. For example, Random Walk based approaches \cite{zhou2005text,szummer2001partially} mainly focus on using distance or angle between node vectors to calculate transition probability while ignoring the information represented by node vectors. Secondly, since the traditional graph-based algorithms are only suitable for specific tasks, there is no unified learning framework for addressing various practical tasks. For example, 
\cite{kaur2018domain} proposes a graph clustering method that requires a domain knowledge-based ontology graph. Lastly, the traditional graph based methods are comparative time inefficient like the Graph Edit Distance-based graph matching methods have exponential time complexity \cite{silva2014bog}.

\subsection{Foundations of GNN}
\hfill\\
To tackle the limitation of traditional graph-based algorithms and better represent non-Euclidean relations in practical applications, Graph Neural Networks are proposed by \cite{GNN}. GNNs have a unified graph-based framework and simultaneously model the graph structure, node, and edge representations. This section will provide the general mathematical definitions of Graph Neural Networks. The general forward process of GNN can be summarised as follows:
\begin{equation}
  \textbf{\textit{H}}^{(l)} =  \mathcal{F}(\textbf{\textit{A}},\textbf{\textit{H}}^{(l-1)})
\end{equation}

\hfill\\
where $\textbf{\textit{A}} \in \mathbb{R}^{n \times n}$ represents the weighted adjacent matrix and $\textbf{\textit{H}}^{(l)} \in \mathbb{R}^{n \times d}$ is the updated node representations at the $l$-th GNN layers by feeding $l-1$-th layer node features $\textbf{\textit{H}}^{(l-1)} \in \mathbb{R}^{n \times k}$ ( $k$ is the dimensions of previous layers node representations ) into pre-defined graph filters $\mathcal{F}$. 
\hfill\\
The most commonly used graph filtering method is defined as follows:
\begin{equation}
  \textbf{\textit{H}}^{(l)} =  \phi(\tilde{\textbf{\textit{A}}}\textbf{\textit{H}}^{(l-1)}\textbf{\textit{W}})
\end{equation}
where $\tilde{\textbf{\textit{A}}} = \textbf{\textit{D}}^{-\frac{1}{2}}\textbf{\textit{AD}}^{-\frac{1}{2}}$ is the normalized symmetric adjacency matrix. $\textbf{\textit{A}} \in \mathbb{R}^{n \times n}$ is the adjacent matrix of graph $G$ and $\textbf{\textit{D}}$ is the degree matrix of $\textbf{\textit{A}}$, where $D_{ii} = \Sigma_{j}A_{ij}$. $\textbf{\textit{W}} \in \mathbb{R}^{k \times d}$ is the weight matrix and $\phi$ is the activation function. If we design a two layers of GNNs based on the above filter could get a vanilla Graph Convolutional Network (GCN) \cite{GCN} framework for text classification:
\begin{equation}
  \textbf{\textit{Y}} =  softmax(\tilde{\textbf{\textit{A}}}(ReLU(\tilde{\textbf{\textit{A}}}\textbf{\textit{H}}\textbf{\textit{W}}^{(0)}))\textbf{\textit{W}}^{(1)})
\end{equation}
where $\textbf{\textit{W}}^0$ and $\textbf{\textit{W}}^1$ represent different weight matrix for different GCN layers and $\textbf{\textit{H}}$ is the input node features. $ReLU$ function is used for non-linearization and $softmax$ is used to generated predicted categories $\textbf{\textit{Y}}$. 

\subsection{GNN for Text Classification}
In this paper, we mainly discuss how GNNs are applied in Text Classification tasks. Before we present the specific applications in this area, we first introduce the key concepts of applying GNNs to text classification from a designer's view. We suppose for addressing a text classification task need to design a graph $G = (V,E)$. The general procedures include \textit{Graph Construction}, \textit{Initial Node Representation}, \textit{Edge Representations}, \textit{Training Setup}.
\subsubsection{Graph Construction} 
\hfill\\
Some applications have explicit graph structures including constituency or dependency graphs \cite{IGCN2020}, knowledge graphs \cite{ostendorff2019enriching, marin2014learning}, social networks \cite{GFN} without constructing graph structure and defining corresponding nodes and edges. However, for text classification, the most common graph structures are implicit, which means we need to define a new graph structure for a specific task such as designing a word-word or word-document co-occurrence graph. In addition, for text classification tasks, the graph structure can be generally classified into two types: 
\begin{itemize}
\item \textbf{\textit{Corpus-level}}/\textbf{\textit{Document-level}}.  
Corpus-level graphs intend to construct the graph to represent the whole corpus such as \cite{TEXTGCN2019,TENSORGCN2020, BERTGCN2021, SGC2019},
while the document-level graphs focus on representing the non-Euclidean relations existing in a single text body like \cite{IDGL2020, MPAD2020, TEXTING2020}. Supposing a specific corpus $\mathcal{C}$ contains a set of documents (text bodies) $\mathcal{C} = \{D_1,D_2,...,D_j\}$ and each $D_i$ contains a set of tokens $D_i = \{t_{i_1},t_{i_2},...,t_{i_k}\}$. The vocabulary of $\mathcal{C}$ can be represented as $\mathcal{D}  = \{t_1,t_2,...,t_l\}$, where $l$ is the length of $\mathcal{D}$. For the most commonly adopted corpus-level graph $G_{corpus} = (V_{corpus},E_{corpus})$, a node $v_i$ in $V_{corpus}$ follows $v_i \in \mathcal{C} \cup \mathcal{D}$ and the edge $e_{ij} \in E_{corpus}$ is one kind of relations between $v_i$ and $v_j$. Regarding the document level graph $G_{doc_i} = (V_{doc_i},E_{doc_i})$, a node $v_{i_j}$ in $V_{doc_i}$ follows $v_{i_j} \in D_i$.
\end{itemize}
After designing the graph-scale for the specific tasks, specifying the graph types is also important to determine the nodes and their relations. For text classification tasks, the commonly used graph construction ways can be summarized into:
\begin{itemize}
    \item \textbf{\textit{Homogeneous}}/\textbf{\textit{Heterogeneous Graphs}}: homogeneous graphs have the same node and edge type while heterogeneous graphs have various node and edge types. For a graph $G = (V,E)$, we use $\mathcal{N}^v$ and $\mathcal{N}^e$ to represent the number of types of $V$ and $E$. If $\mathcal{N}^v = \mathcal{N}^e = 1$, $G$ is a homogeneous graph. If $\mathcal{N}^v >1 $ or $ \mathcal{N}^e > 1$, $G$ is a heterogeous graph.
   \item \textbf{\textit{Static}}/\textbf{\textit{Dynamic Graphs}}: Static graphs aim to use the constructed graph structure by various external or internal information to leverage to enhance the initial node representation such as dependency or constituency graph \cite{IGCN2020}, co-occurrence between word nodes \cite{TEXTING2020}, TF-IDF between word and document nodes \cite{TEXTGCN2019, SGC2019, NMGC2021} and so on. However, compared with the static graph, the dynamic graph initial representations or graph topology are changing during training without certain domain knowledge and human efforts. The feature representations or graph structure can jointly learn with downstream tasks to be optimised together. For example, \cite{DHTG2020} proposed a novel topic-awared GNN text classification model with dynamically updated edges between topic nodes with others (e.g. document, word). Piao et al. \cite{TEXTSSL2021} also designed a dynamic edge based graph to update the contextual dependencies between nodes. Additionally, \cite{IDGL2020} propose a dynamic GNN model to jointly update the edge and node representation simultaneously. We provide more details about above mentioned models in Section~\ref{sec:corpussec} and Section~\ref{sec:docsec}. 
\end{itemize}
Another widely used pair of graph categories are \textbf{\textit{directed}} or \textbf{\textit{undirected}} graphs based on whether the directions of edges are bi-directional or not. For text classification, most of the GNN designs are following the unidirectional way. In addition, those graph type pairs are not parallel which means those types can be combined. 

\subsubsection{Initial Node Representation}
\hfill\\

Based on the pre-defined graph structure and specified graph type, selecting the appropriate initial node representations is the key procedure to ensure the proposed graph structure can effectively learn node. According to the node entity type, the existing node representation approaches for text classification can be generally summarised into:
\begin{itemize}
    \item \textbf{\textit{Word-level Representation}}: non-context word embedding methods such as GloVe \cite{pennington2014GloVe}, Word2vec  \cite{mikolov2013efficient}, FastText \cite{bojanowski2017enriching} are widely adopted by many GNN-based text classification framework to numerically represent the node features. However, those embedding methods are restricted to capturing only syntactic similarity and fail to represent the complex semantic relationships between words, as well as, they cannot capture the meaning of out-of-vocabulary (OOV) words, and their representations are fixed. Therefore, there are some recent studies selecting ELMo \cite{peters-etal-2018-deep}, BERT \cite{devlin2018bert}, GPT \cite{radford2018improving} to get contextual word-level node representation. Notably, even if one-hot encoding is the simplest word representation method, there are many GNN-based text classifiers using one-hot encoding and achieving state-of-the-art performance. Few frameworks use randomly initialised vectors to represent the word-level node features.
    \item \textbf{\textit{Document-level Representation}}: similar to other NLP applications, document level representations are normally acquired by aggregating the word level representation via some deep learning frameworks. For example, some researchers select by extracting the last-hidden state of LSTM or using the [CLS] token from BERT to numerically represent the input text body. Furthermore, it is also a commonly used document-level node representation way to use TF-IDF based document vectors.
\end{itemize}
Most GNN based text classification frameworks will compare the performance between different node representation methods to conduct quantitative analysis, as well as provide reasonable justifications for demonstrating the effectiveness of the selected initial node representation based on defined graph structure. 

\subsubsection{Edge Features}
\hfill\\
Well-defined edge features can effectively improve the graph representation learning efficiency and performance to exploit more explicit and implicit relations between nodes. Based on the predefined graph types, the edge feature types can be divided into \textit{\textbf{structural features}} and \textit{\textbf{non-structural features}}. The structural edge features are acquired from explicit relations between nodes such as dependency or constituency relation between words, word-word adjacency relations, etc. Those relations between nodes are explicitly defined and are also widely employed in other NLP applications. However, more commonly used edge features are non-structural features which implicitly existed between the nodes and specifically applied to specific graph-based frameworks. The typically non-structural edge features are firstly defined by \cite{kim2014convolutional} for GNNs-based text classification tasks including:
\begin{itemize}
    \item \textbf{\textit{PMI}}  measures the co-occurrence between two words in a sliding window $W$ and is calculated as:
\begin{gather}
    \text{PMI}(i,j) = log\frac{p(i,j)}{p(i)p(j)}\\
    p(i,j) = \frac{\#W(i,j)}{\#W}\\
    p(i) = \frac{\#W(i)}{\#W} 
\end{gather}
where $\#W$ is the number of windows in total, and $\#W(i)$, $\#W(i,j)$ shows the number of windows containing word $i$ and both word $i$ and $j$ respectively. 
    \item \textbf{\textit{TF-IDF}} is the broadly used weight of the edges between document-level nodes and word-level nodes. 
    \
\end{itemize}
Except for those two widely used implicit edge features, some specific edge weighting methods are proposed to meet the demands of particular graph structures for exploiting more information of input text bodies. 

\subsubsection{Training Setup}
\hfill\\
After specifying the graph structure and types, the graph representation learning tasks and training settings also need to be determined to decide how to optimise the designed GNNs. Generally, the graph representation learning tasks can be categorised into three levels including \textbf{\textit{Node-level}}, \textbf{\textit{Graph-level}} and \textbf{\textit{Edge-level}}. Node-level and Graph-level tasks involve node or graph classification, clustering, regression, etc, while Edge-level tasks include link prediction or edge classification for predicting the relation existence between two nodes or the corresponding edge categories. 

Similar to other deep learning model training settings, GNNs also can be divided into \textbf{\textit{supervised}}, \textbf{\textit{semi-supervised}} and \textbf{\textit{unsupervised training settings}}. Supervised training provides labelled training data, while unsupervised training utilises unlabeled data to train the GNNs. However, compared with supervised or unsupervised learning, semi-supervised learning methods are broadly used by GNNs designed for text classification applications which could be classified into two types:
\begin{itemize}
    \item \textbf{\textit{Inductive Learning}} adjusts the weights of proposed GNNs based on a labelled training set for learning the overall statistics to induce the general trained model for following processing. The unlabeled set can be fed into the trained GNNs to compute the expected outputs. 
    \item \textbf{\textit{Transductive Learning}} intends to exploit labelled and unlabeled sets simultaneously for leveraging the relations between different samples to improve the overall performance.
\end{itemize}


\begin{table}[t]
\caption{Commonly used notations in Graph Neural Networks}
\label{tab:notations}
\centering
\begin{tabular} {  l l p{7cm} } \toprule
        \textbf{Notations}& \textbf{Descriptions} \\ \midrule
        $G$& A graph. \\ \hline
        $V$& The set of nodes in a graph.\\ \hline
        $E$& The set of edges in a graph.\\ \hline
        $e_{ij}$ & An edge between node $i$ and node $j$.\\ \hline
        $N_i$ & The neighbors of a node $i$. \\ \hline
        $\bm{A}$ & The graph adjacency matrix.  \\ \hline
        $\tilde{\bm{A}}$ & The normalized matrix
        $\bm{A}$. \\ \hline
        $\tilde{\bm{A}}^k, k\in Z$ & The $k^{th}$ power of
        $\tilde{\bm{A}}$. \\ \hline
        $[\bm{A}||\bm{B}]$ & The concatenation of $\bm{A}$ and $\bm{B}$. \\ \hline
        $\bm{D}$ & The degree matrix of $\bm{A}$. $\bm{D}_{ii} = \Sigma_{j=1}^n \bm{A}_{ij}$. \\ \hline

        $\bm{W^{(l)}}$ & The weight matrix of layer $l$. \\ \hline
        
        $\bm{H} \in \bm{R}^{n\times d}$ & The feature matrix of a graph. \\ \hline
        $\bm{H^{(l)}} \in \bm{R}^{n\times d}$ & The feature matrix of a graph at layer $l$. \\ \hline
        $\bm{h_i} \in \bm{R}^n$ & The feature vector of the node $i$ \\ \hline
            $\bm{h_i^{(l)}} \in \bm{R}^n$ & The feature vector of the node $i$ at layer $l$.\\ \hline

        $\bm{Z} \in \bm{R}^{n\times d}$ & The output feature matrix of a graph. \\ \hline

        $\bm{z_i} \in \bm{R}^n$ & The output feature vector of the node $i$ \\
         
        \bottomrule
    \end{tabular}
\end{table}

\section{Corpus-level GNN for Text Classification}\label{sec:corpussec}
We define a corpus-level Graph Neural Network as ``constructing a graph to represent the whole corpus", thus, only one or several graphs will be built for the given corpus. We categorize Corpus-level GNN into four subcategories based on the types of nodes shown in the graph.

\subsection{Document and Word Nodes as a Graph}
Most corpus-level graphs include word nodes and document nodes and there are word-document edges and word-word edges.  By applying $K$(normally $K$=2 or 3) layer GNN, word nodes will serve as a bridge to propagate the information from one document node to another.

\subsubsection{PMI and TF-IDF as graph edges: TextGCN, SGC, S$^2$GC, NMGC, TG-Transformer, BertGCN}
\hfill

\textbf{TextGCN}\cite{TEXTGCN2019}\textbf{ }
\citet{TEXTGCN2019} builds a corpus-level graph with training document nodes, test document nodes and word nodes. Before constructing the graph, a common preprocessing method\cite{kim2014convolutional} has been applied and words shown fewer than 5 times or in NLTK\cite{bird2009natural} stopwords list have been removed.
The edge value between the document node and the word node is TF-IDF and that between the word nodes is PMI. The adjacency matrix of this graph shows as follows.
\begin{equation}
  A_{ij} = 
  \begin{cases}
    \text{PMI}(i,j)&i, j\text{ are words}, \text{PMI}(i,j)>0\\
    \text{TF-IDF}_{i,j}&i\text{ is document}, j \text{ is word}\\
    1&i=j\\
    0&\text{otherwise}
  \end{cases}
\end{equation}
A two-layer GCN is applied to the graph, and the dimension of the second layer output equals to the number of classes in the dataset. Formally, the forward propagation of TextGCN shows as:
\begin{equation}
  \bm{Z} = \text{softmax}(\tilde{\bm{A}}(\text{ReLU}(\tilde{\bm{A}}\bm{X}\bm{W}^{(0)}))\bm{W}^{(1)})
\end{equation}
where $\tilde{A}$ is the normalized adjacency of $A$ and $X$ is one-hot embedding. $W_0$ and $W_1$ are learnable parameters of the model. The representation on training documents is used to calculate the loss and that on test documents is for prediction. TextGCN is the first work that treats a text classification task as a node classification problem by constructing a corpus-level graph and has inspired many following works.

Based on TextGCN, several works follow the same graph construction method and node initialization but apply different graph propagation models.

\textbf{SGC}\cite{SGC2019}\textbf{ }
To make GCN efficient, SGC (Simple Graph Convolution) removes the nonlinear activation function in GCN layers, therefore, the K-layer propagation of SGC shows as:
\begin{equation}
  \bm{Z} =  \text{softmax}(\tilde{\bm{A}}...(\tilde{\bm{A}}(\tilde{\bm{A}}\bm{X}\bm{W}^{(0)})\bm{W}^{(1)})...\bm{W}^{(K)})
\end{equation}
which can be reparameterized into
\begin{equation}
  \bm{Z} =  \text{softmax}(\tilde{\bm{A}}^K\bm{X}\bm{W})
\end{equation}
and $K$ is 2 when applied to text classification tasks. With a smaller number of parameters and only one feedforward layer, SGC saves computation time and resources while improving performance.

\textbf{S$^2$GC}\cite{S2GCN2020}\textbf{ }
To solve the oversmoothing issues in GCN, \citet{S2GCN2020} propose Simple Spectral Graph Convolution(S$^2$GC) which includes self-loops using Markov Diffusion Kernel. The output of S$^2$GC is calculated as:
\begin{equation}
  \bm{Z} =  \text{softmax}(\frac{1}{K}\Sigma_{k=0}^K\tilde{\bm{A}}^k\bm{X}\bm{W})
\end{equation}
and can be generalized into:
\begin{equation}
  \bm{Z} =  \text{softmax}(\frac{1}{K}\Sigma_{k=0}^K((1-\alpha)\tilde{\bm{A}}^k\bm{X}+\alpha\bm{X})\bm{W})
\end{equation}
Similarly, $K$ = 2 on text classification tasks and $\alpha$ denotes the trade-off between self-information of the node and consecutive neighbourhood information. S$^2$GC can also be viewed as introducing skip connections into GCN.

\textbf{NMGC}\cite{NMGC2021}\textbf{ }
Other than using the sum of each GCN layer in S$^2$GC, NMGC applies min pooling using the Multi-hop neighbour Information Fusion (MIF) operator to address oversmoothing problems. A MIF function is defined as:
\begin{equation}
  \text{MIF}(K) = \text{min}(\tilde{\bm{A}}\bm{X}\bm{W},\tilde{\bm{A}}^2\bm{X}\bm{W},...,\tilde{\bm{A}}^K\bm{X}\bm{W})
\end{equation}
NMGC-K firstly applies a MIF($K$) layer then a GCN layer and K is 2 or 3. For example, when $K$ = 3, the output is:
\begin{equation}
  \bm{Z} =  \text{softmax}(\tilde{\bm{A}}(\text{ReLU min}(\tilde{\bm{A}}\bm{X}\bm{W}^{(0)},\tilde{\bm{A}}^2\bm{X}\bm{W}^{(0)},\tilde{\bm{A}}^3\bm{X}\bm{W}^{(0)}))\bm{W}^{(1)})
\end{equation}
NMGC can also be treated as a skip-connection in Graph Neural Networks which makes the shallow layer of GNN contribute to the final representation directly.

\textbf{TG-Transformer}\cite{TGTRANSFORMER2020}\textbf{ }
TextGCN treats the document nodes and word nodes as the same type of nodes during propagation, and to introduce heterogeneity into the TextGCN graph, TG-Transformer (Text Graph Transformer) adopts two sets of weights for document nodes and word nodes respectively. To cope with a large corpus graph, subgraphs are sampled from the TextGCN graph using PageRank algorithm\cite{page1999pagerank}. The input embedding of is the sum of three types of embedding: pretrained GloVe embedding, node type embedding, and Weisfeiler-Lehman structural encoding\cite{niepert2016learning}. During propagation, self-attention\cite{vaswani2017attention} with graph residual\cite{zhang2019gresnet} is applied.

\textbf{BertGCN}\cite{BERTGCN2021}\textbf{ }
To combine BERT\cite{BERT} and TextGCN, BertGCN enhances TextGCN by replacing the document node initialization with the BERT [CLS] output of each epoch and replacing the word input vector with zeros. BertGCN trains BERT and TextGCN jointly by interpolating the output of TextGCN and BERT:
\begin{equation}
  \bm{Z} = \lambda\bm{Z}_{GCN} + (1-\lambda)\bm{Z}_{BERT}
\end{equation}
where $\lambda$ is the trade-off factor. To optimize the memory during training, a memory bank is used to track the document input and a smaller learning rate is set to BERT module to remain the consistency of the memory bank. BertGCN shows that with the help of TextGCN, BERT can achieve better performance.

\subsubsection{Multi-Graphs/Multi-Dimensional Edges: TensorGCN, ME-GCN}
\hfill

\textbf{TensorGCN}\cite{TENSORGCN2020}\textbf{ }
Instead of constructing a single corpus-level graph, TensorGCN builds three independent graphs: Semantic-based graph, Syntactic-based graph, and Sequential-based graph to incorporate semantic, syntactic and sequential information respectively and combines them into a tensor graph.

Three graphs share the same set of TF-IDF values for the word-document edge but different values for word-word edges. Semantic-based graph extracts the semantic features from a trained Long short-term memory(LSTM)\cite{hochreiter1997long} model and connects the words sharing high similarity. Syntactic-based graph uses Stanford CoreNLP parser\cite{stanfordcorenlp} and constructs edges between words when they have a larger probability of having dependency relation. For Sequential-based graph, PMI value is applied 
as TextGCN does.

The propagation includes intra-graph propagation and inter-graph propagation. The model first applies the GCN layer on three graphs separately as intra-graph propagation. Then the same nodes on three graphs are treated as a virtual graph and another GCN layer is applied as inter-graph propagation. 

\textbf{ME-GCN}\cite{MEGCN2022}\textbf{ }
To fully utilize the corpus information and analyze rich relational information of the graph, ME-GCN (Multi-dimensional Edge-Embedded GCN) builds a graph with multi-dimensional word-word, word-document and document-document edges. Word2vec and Doc2vec embedding is firstly trained on the given corpus and the similarity of each dimension of trained embedding is used to construct the multi-dimensional edges. The trained embedding also serves as the input embedding of the graph nodes. During propagation, GCN is firstly applied on each dimension and representations on different dimensions are either concatenated or fed into a pooling method to get the final representations of each layer.

\subsubsection{Making TextGCN Inductive: HeteGCN, InducT-GCN, T-VGAE}
\hfill

\textbf{HeteGCN}\cite{HETEGCN2021}\textbf{ }
HeteGCN (Heterogeneous GCN) optimizes the TextGCN by decomposing the TextGCN undirected graph into several directed subgraphs. Several subgraphs from TextGCN graph are combined sequentially as different layers: feature graph (word-word graph), feature-document graph (word-document graph), and document-feature graph (document-word graph). Different combinations were tested and the best model is shown as:
\begin{gather}
  \bm{Z} =  \text{softmax}(\bm{A}_{w-d}(\text{ReLU}(\bm{A}_{w-w}\bm{X}_{w}\bm{W}^{(0)}))\bm{W}^{(1)})
\end{gather}
where $\bm{A}_{w-w}$ and $\bm{A}_{w-d}$ show the adjacency matrix for the word-word subgraph and word-document subgraph. Since the input of HeteGCN is the word node embeddings without using document nodes, it can also work in an inductive way while the previous corpus-level graph text classification models are all transductive models.

\textbf{InducT-GCN}\cite{wang2022induct}\textbf{ }
InducT-GCN (InducTive Text GCN) aims to extend the transductive TextGCN into an inductive model. Instead of using the whole corpus for building the graph, InducT-GCN builds a training corpus graph and makes the input embedding of the document as the TF-IDF vectors, which aligns with the one-hot word embeddings. The weights are learned following TextGCN but InducT-GCN builds virtual subgraphs for prediction on new test documents. 

\textbf{T-VGAE}\cite{TVGAE2021}\textbf{ }
T-VGAE (Topic Variational Graph Auto-Encoder) applies Variational Graph Auto-Encoder on the latent topic of each document to make the model inductive. A vocabulary graph $A_v$ which connects the words using PMI values is constructed while each document is represented using the TF-IDF vector. All the document vectors are stacked into a matrix which can also be treated as a bipartite graph $A_d$.
Two graph auto-encoder models are applied on $A_v$ and $A_d$ respectively. The overall workflow shows as:
\begin{gather}
  \bm{Z}_v = \text{Encoder}_{GCN}(\bm{A}_v,\bm{X}_v) \\
  \bm{Z}_d = \text{Encoder}_{UDMP}(\bm{A}_d,\bm{Z}_v)\\
  \bm{A}_v^* = \text{Decoder}(\bm{Z}_v)\\
  \bm{A}_d^* = \text{Decoder}(\bm{Z}_d,\bm{Z}_v)
\end{gather}
where $X^v$ is an Identity Matrix. The $\text{Encoder}_{GCN}$ and the decoders are applied following $VGAE$\cite{kipf2016variational} while $\text{Encoder}_{UDMP}$ is an unidirectional message passing variant of $\text{Encoder}_{GCN}$. The training objective is minimising the reconstruction error and $Z_d$ is used for the classification task.

\subsection{Document Nodes as a Graph}
To show the global structure of the corpus directly, some models only adopt document nodes in the non-heterogeneous graph.

\textbf{knn-GCN}\cite{benamira2019semi}\textbf{ }
knn-GCN constructs a k–nearest-neighbours graph by connecting the documents with their $K$ nearest neighbours using Euclidean distances of the embedding of each document. The embedding is generated in an unsupervised way: either using the mean of pretrained GloVe word vectors or applying LDA\cite{blei2003latent}. Both GCN and Attention-based GNN\cite{thekumparampil2018attention} are used as the graph model.

\textbf{TextGTL}\cite{TEXTGTL2021}\textbf{ }
Similar to TensorGCN, TextGTL (Text-oriented Graph-based Transductive Learning) constructs three different document graphs: Semantics Text Graph, Syntax Text Graph, and Context Text Graph while all the graphs are non-heterogeneous. Semantics Text Graph uses Generalized Canonical Correlation Analysis\cite{bach2002kernel} and trains a classifier to determine the edge values between two document nodes. Syntax Text Graph uses the Stanford CoreNLP dependency parser\cite{stanfordcorenlp} to construct units and also trains a classifier. Context Text Graph defines the edge values by summing up the PMI values of the overlapping words in two documents.
Two GCN layers are applied and the output of each graph is mixed as the output of this layer and input for the next layer for all three graphs:
\begin{gather}
  \bm{H}^{(1)} = \sigma(\bm{A}\bm{H}^{(0)}\bm{W}^{(0)})\\
  \bm{H}^{(2)} = \sigma(\bm{A}[\bm{H}_{sem}^{(1)}||\bm{H}_{syn}^{(1)}||\bm{H}_{seq}^{(1)}]\bm{W}^{(1)})\\
  \bm{Z} = \text{Pooling}_{mean}(\bm{H}_{sem}^{(2)},\bm{H}_{syn}^{(2)},\bm{H}_{seq}^{(2)})
\end{gather}
where $H^{(0)}$ is the TF-IDF vector of the documents. Data augmentation with super nodes is also applied in TextGTL to strengthen the information in graph models.

\subsection{Word Nodes as a Graph}
By neglecting the document nodes in the graph, a graph with only word nodes shows good performance in deriving the graph-based embedding and is used for downstream tasks. Since no document nodes are included, this method can be easily adapted as an inductive learning model.

\textbf{VGCN-BERT}\cite{VGCNBERT2020}\textbf{ }
VGCN-BERT enhances the input embedding of BERT by concatenating it with the graph embedding. It first constructs a vocabulary graph and uses PMI as the edge value. A variant of the GCN layer called VGCN(Vocabulary GCN) is applied to derive the graph word embedding:
\begin{equation}
  \bm{X}_{Graph} = \text{ReLU}(\bm{X}_{BERT}\bm{A}\bm{W}^{(0)})\bm{W}^{(1)}
\end{equation}
where BERT embedding is used as the input. The graph word embeddings are concatenated with BERT embedding and fed into the BERT as extra information.

\subsection{Extra Topic Nodes in the Graph}
Topic information of each document can also provide extra information in corpus-level graph neural networks. Several models also include topic nodes in the graph.
\subsubsection{Single Layer Topic nodes: HGAT, STGCN}
\hfill

\textbf{HGAT}\cite{HGAT2019}\textbf{ }
HGAT (Heterogeneous GAT) applies LDA\cite{blei2003latent} to extract topic information for each document, top $P$ topics with the largest probabilities are selected as connected with the document. Instead of using the words directly, to utilize the external knowledge HGAT applies the entity linking tool TAGME\footnote{https://sobigdata.d4science.org/group/tagme/} to identify the entities in the document and connects them. The semantic similarity between entities using pretrained Word2vec with threshold is used to define the connectedness between entity nodes. Since the graph is a heterogeneous graph, a HIN (heterogeneous information network) model is implemented which propagates solely on each sub-graphs depending on the type of node. An HGAT model is applied by considering type-level attention and node-level attention. For a given node, the type-level attention learns the weights of different types of neighbouring nodes while node-level attention captures the importance of different neighbouring nodes when ignoring the type. By using the dual attention mechanism, HGAT can capture the information of type and node at the same time.

\textbf{STGCN}\cite{yan2013biterm}\textbf{ }
In terms of short text classification, STGCN (Short-Text GCN) applies BTM to get topic information to avoid the data sparsity problem from LDA. The graph is constructed following TextGCN while extra topic nodes are included. The edge values of word-topic and document-topic are from BTM and a classical two-layer GCN is applied. The word embeddings learned from STGCN are concatenated with BERT embeddings and a bi-LSTM model is applied for final prediction.

\subsubsection{Multi-layer Topic Nodes: DHTG}
\hfill

\textbf{DHTG}\cite{DHTG2020}\textbf{ }
To capture different levels of information, DHTG (Dynamic Hierarchical Topic Graph) introduces hierarchical topic-level nodes in the graph from fine-grain to coarse. Poisson gamma belief network (PGBN)\cite{DBLP:conf/nips/ZhouCC15} is used as a probabilistic deep topic model. The first-layer topics are from the combination of words, while deeper layers are generated by previous layers' topics with the weights of PGBN, and the weights serve as the edge values of each layer of topics. For the topics on the same layer, the cosine similarity is chosen as the edge value. A two-layer GCN is applied and the model is learned jointly with PGBN, which makes the edge of the topics dynamic.

\subsection{Critical Analysis}
Compared with sequential models like CNN and LSTM, corpus-level GNN is able to capture the global corpus structure information with word nodes as bridges between document nodes and shows great performance without using external resources like pretrained embedding or pretrained model. However, the improvement in performance is marginal when pretrained embedding is included. Another issue is that most corpus-level GNN is transductive learning which is not applicable in the real world. Meanwhile, constructing the whole corpus into a graph requires large memory space especially when the dataset is large.

A detailed comparison of corpus-level GNN is displayed in Table \ref{tab:detail_comparison}.

\section{Document-level GNN for Text Classification}\label{sec:docsec}
By constructing the graph based on each document, a graph classification model can be used as a text classification model. Since each document is represented by one graph and new graphs can be built for test documents, the model can easily work in an inductive way.

\subsection{Local Word Consecutive Graph}
The simplest way to convert a document into a graph with words as nodes is by connecting the consecutive words within a sliding window.
\subsubsection{Simple consecutive graph models: Text-Level-GNN, MPAD, TextING}
\hfill

\textbf{Text-Level-GNN}\cite{TEXTLEVELGNN2019}\textbf{ }
Text-Level-GNN applies a small sliding window and constructs the graph with a small number of nodes and edges in each graph, which saves memory and computation time. The edge value is trainable and shared across the graphs when connecting the same two words, which also brings global information.

Unlike corpus-level graph models, Text-Level-GNN applies a message passing mechanism (MPM)\cite{gilmer2017neural} instead of GCN for graph learning. For each node, the neighbour information is aggregated using max-pooling with trainable edge values as the AGGREGATE function and then the weighted sum is used as the COMBINE function. To get the representation of each graph, sum-pooling and an MLP classifier are applied as the READOUT function. The propagation shows as:
\begin{gather}
     \bm{h}^{(l+1)}_i = (1-\alpha)(max_{n\in \mathcal{N}_i}e_{ni}\bm{h}^{(l)}_n)+ \alpha\bm{h}^{(l)}_i \\
  \bm{z_i} = \text{softmax}(\bm{W}\Sigma_i\bm{h}_i+\bm{b})
\end{gather}
where $\bm{h}^{(l)}_i$ is $i$th word node presentation of layer $l$, $e_{ni}$ is edge weight from node $n$ to node $i$. A two-layer MPM is applied, and the input of each graph is pretrained GloVe vectors.

\textbf{MPAD}\cite{MPAD2020}\textbf{ }
MPAD (Message Passing Attention Networks) connects words within a sliding window of size 2 but also includes an additional master node connecting all nodes in the graph. The edge only shows the connectedness of each pair of word nodes and is fixed. A variant of Gated Graph Neural Networks is applied where the AGGREGATE function is the weighted sum and the COMBINE function is GRU\cite{chung2014empirical}. Self-attention is applied in the READOUT function.

To learn the high-level information, the master node is directly concatenated with the READOUT output, working as a skip connection mechanism. To get the final representation, each layer's READOUT results are concatenated to capture multi-granularity information. Pretrained Word2vec is used as the initialization of word nodes input.

\textbf{TextING}\cite{TEXTING2020}\textbf{ }
To simplify MPAD, TextING ignores the master node in the document-level graphs, which makes the graph sparser. Compared with Text-Level-GNN, TextING remains fixed edges. A similar AGGREGATE and COMBINE function are applied under the concept of e Gated Graph
Neural Networks(GGNN)\cite{DBLP:journals/corr/LiTBZ15} with the weighted sum and GRU. However, for the READOUT function, soft attention is used and both max-pooling and mean-pooling are applied to make sure that "every word plays a role in the text and the keywords should contribute more explicitly".

\subsubsection{Advanced graph models: MLGNN, TextSSL, DADGNN}
\hfill

\textbf{MLGNN}\cite{MLGNN2021}\textbf{ }
MLGNN (Multi-level GNN) builds the same graph as TextING but introduces three levels of MPM: bottom-level, middle-level and top-level. In the bottom-level MPM, the same method with Text-Level-GNN is applied with pretrained Word2vec as input embedding but the edge is non-trainable. In the middle level, a larger window size is adopted and Graph Attention Networks(GAT)\cite{velickovic2018graph} is applied to learn long distant word nodes information. In the top-level MPM, all word nodes are connected and multi-head self-attention\cite{vaswani2017attention} is applied. By applying three different levels of MPM, MLGNN learns multi-granularity information well. 

\textbf{DADGNN}\cite{DADGNN2021}\textbf{ }
DADGNN (Deep Attention Diffusion GNN) constructs the same graph as TextING but uses attention diffusion to overcome the oversmoothing issue. Pretrained word embedding is used as the input of each node and an MLP layer is applied. Then, the graph attention matrix is calculated based on the attention to the hidden states of each node. The diffusion matrix is calculated as 
\begin{equation}
  \bm{T} = \Sigma_{n=0}^{\infty}\epsilon_n\bm{A}^n
\end{equation}
where $A$ is the graph attention matrix and $\epsilon$ is the learnable coefficients. $A^n$ plays a role of connecting $n$-hop neighbours and \citet{DADGNN2021} uses $n \in [4,7]$ in practice. A multi-head diffusion matrix is applied for layer propagation. 

\textbf{TextSSL}\cite{TEXTSSL2021}\textbf{ }
To solve the word ambiguity problem and show the word synonymity and dynamic contextual dependency, TextSSL (Sparse Structure Learning) learns the graph using intra-sentence neighbours and inter-sentence neighbours simultaneously. The local syntactic neighbour is defined as the consecutive words and trainable edges across graphs are also included by using Gumbel-softmax
. By applying sparse structure learning, TextSSL manages to select edges with dynamic contextual dependencies.

\subsection{Global Word Co-occurrence Graph}
Similar to the TextGCN graph, document-level graphs can also use PMI as the word-word edge values.
\subsubsection{Only global word co-occurrence: DAGNN}
\hfill

\textbf{DAGNN}\cite{DAGNN2019}\textbf{ }
To address the long-distance dependency, hierarchical information and cross-domain learning challenges in domain-adversarial text classification tasks, \citet{DAGNN2019} propose DAGNN (Domain-Adversarial Graph Neural Network). Each document is represented by a graph with content words as nodes and PMI values as edge values, which can capture long-distance dependency information. Pretrained FastText is chosen as the input word embeddings to handle the out-of-vocabulary issue and a GCN model with skip connection is used to address the oversmoothing problem. The propagation is formulated as:
\begin{gather}
  \bm{H}^{(l+1)} = (1-\alpha)\tilde{\bm{A}}\bm{H}^{(l)} + \alpha\bm{H}^{(0)}
\end{gather}
To learn the hierarchical information of documents, DiffPool\cite{ying2018hierarchical} is applied to assign each document into a set of clusters. Finally, adversarial training is used to minimize the loss on source tasks and maximize the differentiation between source and target tasks.

\subsubsection{Combine with Extra Edges: ReGNN, GFN}
\hfill

\textbf{ReGNN}\cite{REGNN2019}\textbf{ }
To capture both global and local information, ReGNN (Recursive Graphical Neural Network) uses PMI together with consecutive words as the word edges. And graph propagation function is the same as GGNN while additive attention\cite{DBLP:journals/corr/BahdanauCB14} is applied in aggregation. Pretrained GloVe is the input embedding of each word node.
 
\textbf{GFN}\cite{GFN}\textbf{ }
GFN (Graph Fusion Network) builds four types of graphs using the word co-occurrence statistics, PMI, the similarity of pretrained embedding and Euclidean distance of pretrained embedding. Although four corpus-level graphs are built, the graph learning actually happens on subgraphs of each document, making the method a document-level GNN. For each subgraph, each type of graph is learned separately using the graph convolutional method and then a fusion method of concatenation is used. After an MLP layer, average pooling is applied to get the document representation.

\subsection{Other word graphs}
Some other ways of connecting words in a document have been explored.

\textbf{HyperGAT}\cite{HYPERGAT2020}\textbf{ }
\citet{HYPERGAT2020} proposes HyperGAT (Hypergraph Attention Networks) which builds hypergraphs for each document to capture high-level interaction between words. Two types of hyperedges are included: sequential hyperedges connecting all words in a sentence and semantic hyperedges connecting top-K words after getting the topic of each word using LDA. Like traditional hypergraph propagations, HyperGAT follows the same two steps of updating but with an attention mechanism to highlight the key information: Node-level attention is applied to learn hyperedges representations and edge-level attention is used for updating node representations.

\textbf{IGCN}\cite{IGCN2020} \textbf{ }
Contextual dependency helps in understanding a document and the graph neural network is no exception. IGCN constructs the graph with the dependency graph to show the connectedness of each pair of words in a document. Then, the word representation learned from Bi-LSTM using POS embedding and word embedding is used to calculate the similarity between each pair of nodes. Attention is used for the output to find the important relevant semantic features.

\textbf{GTNT}\cite{GTNT2021}\textbf{ }
Words with higher TF-IDF values should connect to more word nodes, with this in mind, GTNT(Graph Transformer Networks based Text representation) uses sorted TF-IDF value to determine the degree of each node and applies the Havel-Hakimi algorithm\cite{hakami1962realizability} to determine the edges between word nodes. A variant of GAT is applied during model learning. Despite the fact that GAT's attention score is mutual for two nodes, GTNT uses relevant importance to adjust the attention score from one node to another. Pretrained Word2vec is applied as the input of each node.

\subsection{Critical Analysis}
Most document-level GNNs connect consecutive words as edges in the graph and apply a graph neural network model, which makes them similar to CNN where the receptive field enlarges when graph models go deeper. Also, the major differences among document-level GNNs are the details of graph models, e.g. different pooling methods, and different attention calculations, which diminishes the impact of the contribution of these works. Compared with corpus-level GNN, document-level GNN adopts more complex graph models and also suffers from the out-of-memory issue when the number of words in a document is large.

A detailed comparison of document-level GNN is displayed in Table \ref{tab:detail_comparison}.

\begin{table}[]
\caption{Models Detailed Comparison on whether using external resources, how to construct the edge and node input, and whether transductive learning or inductive learning. GloVe and Word2vec are pretrained if not specified. ``emb sim'' is short for ``embedding similarity'', ``dep graph'' is short ``dependency graph''.}
\begin{adjustbox}{width=0.95\textwidth}
\begin{tabular}{c|lllll}
\hline
\textbf{Graph}                            & \textbf{Model}          & \textbf{External Resource}          & \textbf{Edge Construction}                     & \textbf{Node Initialization}                                                 & \textbf{Learning}     \\ \hline
\multirow{17}{*}{\textbf{Corpus-level}}   & TextGCN\cite{TEXTGCN2019}        & N/A                          & pmi, tf–idf                            & one-hot                                                             & transductive \\ \cline{2-6} 
                                 & SGC\cite{SGC2019}            & N/A                          & pmi, tf–idf                            & one-hot                                                             & transductive \\ \cline{2-6} 
                                 & S2GC\cite{S2GCN2020}           & N/A                          & pmi, tf–idf                            & one-hot                                                             & transductive \\ \cline{2-6} 
                                 & NMGC\cite{NMGC2021}           & N/A                          & pmi, tf–idf                            & one-hot                                                             & transductive \\ \cline{2-6} 
                                 & TG-transformer\cite{TGTRANSFORMER2020} & GloVe           & pmi, tf–idf                            & GloVe                                                    & transductive \\ \cline{2-6} 
                                 & BERTGCN\cite{BERTGCN2021}        & BERT                       & pmi, tf–idf                            & doc: 0  word: BERT emb                                              & transductive \\ \cline{2-6} 
                                 & TensorGCN\cite{TENSORGCN2020}      & GloVe, CoreNLP  & emb sim, dep graph, pmi, tf–idf & one-hot                                                             & transductive \\ \cline{2-6} 
                                 & ME-GCN\cite{MEGCN2022}         & N/A                          & emb sim, tf–idf                        & Trained Word2vec/doc2vec                                            & transductive \\ \cline{2-6} 
                                 & HeteGCN\cite{HETEGCN2021}        & N/A                          & pmi, tf–idf                            & one-hot                                                             & inductive    \\ \cline{2-6} 
                                 & InducT-GCN\cite{wang2022induct}     & N/A                          & pmi, tf–idf                            & one-hot, tf–idf vectors                                              & inductive    \\ \cline{2-6} 
                                 & T-VGAE\cite{TVGAE2021}         & N/A                          & pmi                                   & one-hot                                                             & inductive    \\ \cline{2-6} 
                                 & VGCN-BERT\cite{VGCNBERT2020}      & BERT                       & pmi                                   & BERT emb                                                            & transductive \\ \cline{2-6} 
                                 & knn-GCN\cite{benamira2019semi}        & GloVe                      & emb sim                               & GloVe                                                               & transductive \\ \cline{2-6} 
                                 & TextGTL\cite{TEXTGTL2021}        & CoreNLP                    & dep graph, pmi                 & tf–idf vectors                                                       & transductive \\ \cline{2-6} 
                                 & HGAT\cite{HGAT2019}           & TAGME, Word2vec & LDA, entity link, emb sim             & tf-idf, LDA, Word2vec & transductive \\ \cline{2-6} 
                                 & STGCN\cite{STGCN2020}          & BERT                       & pmi, tf–idf, BTM                       & BERT emb                                                            & transductive \\ \cline{2-6} 
                                 & DHTG\cite{DHTG2020}           & N/A                          & PGBN, pmi, tf–idf                      & one-hot                                                             & transductive \\ \hline
\multirow{12}{*}{\textbf{Doc-level}} & Text-Level-GNN\cite{TEXTLEVELGNN2019} & GloVe           & consecutive words                     & GloVe                                                    & inductive    \\ \cline{2-6} 
                                 & MPAD\cite{MPAD2020}           & Word2vec        & consecutive words                     & Word2vec                                                 & inductive    \\ \cline{2-6} 
                                 & TextING\cite{TEXTING2020}        & GloVe           & consecutive words                     & GloVe                                                    & inductive    \\ \cline{2-6} 
                                 & MLGNN\cite{MLGNN2021}          & Word2vec        & consecutive words                     & Word2vec                                                 & inductive    \\ \cline{2-6} 
                                  & DADGNN\cite{DADGNN2021}         & Word2vec/GloVe  & consecutive words                     & Word2vec/GloVe                                           & inductive    \\ \cline{2-6} 
                                 & TextSSL\cite{TEXTSSL2021}        & GloVe           & consecutive words                     & GloVe                                                    & inductive    \\ \cline{2-6} 
                                 & DAGNN\cite{DAGNN2019}          & GloVe           & pmi                                   & GloVe                                                    & inductive    \\ \cline{2-6} 
                                 & ReGNN\cite{REGNN2019}          & GloVe           & consecutive words, pmi                & GloVe                                                    & inductive    \\ \cline{2-6} 
                                 & GFN\cite{GFN}            & GloVe           & pmi, emb sim                          & GloVe                                                    & inductive    \\ \cline{2-6} 
                                 & HyperGAT\cite{HYPERGAT2020}       & N/A                          & LDA, consecutive words                & one-hot                                                             & inductive    \\ \cline{2-6} 
                                 & IGCN\cite{IGCN2020}           & spaCy                      & dep graph                      & LSTM emb                                                            & inductive    \\ \cline{2-6} 
                                 & GTNT\cite{GTNT2021}           & Word2vec/GloVe  & tf–idf sorted value                    & Word2vec/GloVe                                           & inductive    \\ \hline
\end{tabular}
\end{adjustbox}
\label{tab:detail_comparison}
\end{table}

\section{Datasets and Metrics}\label{sec:datasetandmetrics}
\subsection{Datasets}

\begin{table*}[t!]
\begin{center}
   \caption{Commonly Used Text Classification Datasets by GNN-based models}
   \begin{adjustbox}{max width=\textwidth}
  \begin{tabular}{c|p{0.1\textwidth}p{0.17\textwidth}p{0.05\textwidth} p{0.07\textwidth}p{0.07\textwidth}p{0.07\textwidth}p{0.07\textwidth}p{0.07\textwidth}p{0.26\textwidth}}
   \hline
     \multirow{2}{*}{\textbf{Task}} & \multirow{2}{*}{\textbf{Name}} & \multirow{2}{*}{\textbf{Domain}} & \textbf{\#} & \textbf{\# } & \textbf{\# } & \textbf{\# } & \textbf{\# } & \textbf{Ave} & \multirow{2}{*}{\textbf{Models}}\\
      & & & \textbf{Cat.} & \textbf{Docs} & \textbf{Train} & \textbf{Test} & \textbf{Words} & \textbf{Len.} & \textbf{}\\
     \hline
     \multirow{13}{*}{\makecell{\textbf{Topic}\\\textbf{Classification}}} & \multirow{2}{*}{Ohsumed} & \multirow{2}{*}{Bibliography} & \multirow{2}{*}{23} & \multirow{2}{*}{7,400} & \multirow{2}{*}{3,357} & \multirow{2}{*}{4,043} & \multirow{2}{*}{14,157} & \multirow{2}{*}{135.82} & \cite{HGAT2019,DHTG2020,TVGAE2021,TEXTGCN2019,BERTGCN2021,TEXTSSL2021,TENSORGCN2020,NMGC2021,TGTRANSFORMER2020,HYPERGAT2020,TEXTING2020,TEXTGTL2021,REGNN2019}\\\cline{2-10}
     
     & \multirow{2}{*}{R8} & \multirow{2}{*}{News} & \multirow{2}{*}{8} & \multirow{2}{*}{7,674} & \multirow{2}{*}{5,485} & \multirow{2}{*}{2,189} & \multirow{2}{*}{7,688} & \multirow{2}{*}{65.72} & \cite{DADGNN2021,DAGNN2019,TGTRANSFORMER2020,TEXTGCN2019,DHTG2020,TVGAE2021,BERTGCN2021,TEXTSSL2021,TENSORGCN2020,NMGC2021,HGAT2019,HYPERGAT2020,TEXTING2020,TEXTGTL2021,TEXTLEVELGNN2019,REGNN2019}\\
     \cline{2-10}
     
     & \multirow{2}{*}{R52} & \multirow{2}{*}{News} & \multirow{2}{*}{52} & \multirow{2}{*}{9,100} & \multirow{2}{*}{6,532} & \multirow{2}{*}{2,568} & \multirow{2}{*}{8,892} & \multirow{2}{*}{69.82} & \cite{DADGNN2021,DAGNN2019,TGTRANSFORMER2020,TEXTGCN2019,DHTG2020,TVGAE2021,BERTGCN2021,TEXTSSL2021,TENSORGCN2020,NMGC2021,HYPERGAT2020,TEXTING2020,HGAT2019,TEXTLEVELGNN2019,REGNN2019,TEXTGTL2021} \\\cline{2-10}
     
     & \multirow{2}{*}{20NG} & \multirow{2}{*}{News} & \multirow{2}{*}{20} & \multirow{2}{*}{18,846} & \multirow{2}{*}{11,314} & \multirow{2}{*}{7,532} & \multirow{2}{*}{42,757} & \multirow{2}{*}{221.26} & \cite{DHTG2020,DAGNN2019,TVGAE2021,TEXTGCN2019,BERTGCN2021,TEXTSSL2021,TENSORGCN2020,NMGC2021,HGAT2019,HYPERGAT2020,TEXTGTL2021,REGNN2019}\\\cline{2-10}
     
     & AG-News & News & 4 & 127,600 & 120,000 & 7,600 & 128,515 & 44.03 & \cite{MLGNN2021,DADGNN2021,HGAT2019}\\\cline{2-10}
     & WebKB & Web Page & 7 & 4,199 & 2,803 & 1,396 & 7,771 & 133.37 & \cite{DADGNN2021} \\\cline{2-10}
     & TREC & Questions & 6 & 5952 & 5452 & 500 & 9593 & 10.06 &\cite{DADGNN2021}\\\cline{2-10}
     & DBLP & Bibliography & 6 & 81,479 & 61,479 & 20,000 & 25,549 & 8.51 &\cite{DADGNN2021}\\\cline{2-10}
     & DBpedia & Wikipedia & 14 & 630000 & 560000 & 70000 & - & - &\cite{HGAT2019}\\
     \hline
     \multirow{11}{*}{\makecell{\textbf{Sentiment}\\\textbf{Analysis}}} & \multirow{2}{*}{MR} & \multirow{2}{*}{Movie review} & \multirow{2}{*}{2} & \multirow{2}{*}{10,662} & \multirow{2}{*}{7,108} & \multirow{2}{*}{3,554} & \multirow{2}{*}{18,764} & \multirow{2}{*}{20.39} &  \cite{VGCNBERT2020,DHTG2020,TVGAE2021,TEXTGCN2019,DADGNN2021,BERTGCN2021,TEXTSSL2021,TENSORGCN2020,NMGC2021,HGAT2019,HYPERGAT2020,TEXTING2020,TEXTGTL2021,TEXTLEVELGNN2019,REGNN2019,STGCN2020}\\\cline{2-10}
     
     & AAR & Product review & 2 & 3150 & 1575 & 1575 & - & - & \cite{IGCN2020}  \\\cline{2-10}
     & TUA & Airline comments & 2 & 14640 & 7320 & 7320 & - & - & \cite{IGCN2020} \\\cline{2-10}
     & SST-1 & Movie review & 5 & 11,855 & 9,465 & 2,210 & 19,524 & 20.17 & \cite{DADGNN2021} \\\cline{2-10}
     & SST-2 & Movie review & 2 & 9613 & 7,792 & 1,821 & 17539 & 19.67 & \cite{DADGNN2021,VGCNBERT2020}\\\cline{2-10}
     & IMDB & Movie review & 2 & 50,000 & 25,000 & 25,000 & 71,278 & 232.77 & \cite{DADGNN2021,IGCN2020,TGTRANSFORMER2020}\\\cline{2-10}
     
     & Yelp 2014 & Review rating & 5 & 1,125,386 & 900,309 & 225,077 & 476,191 & 148.8 & \cite{TGTRANSFORMER2020}\\\cline{2-10}
     & Twitter & Twitter & 2 & 10000 & - & - & - & - & \cite{HGAT2019}\\\cline{2-10}
     & SenTube-A & Youtube Comments & 2 & 7,400 & 3,357 & 4,043 & 14,157 & 28.54 & \cite{MLGNN2021}\\\cline{2-10}
     & SenTube-T & Youtube Comments & 2 & 6664 & 4997+333 & 1334 & 20,276 & 28.73 & \cite{MLGNN2021} \\\cline{2-10}
     \hline
    \multirow{3}{*}{\textit{\textbf{Other}}} & ArangoHate & Twitter posts & 2 & 7006 & - & - & - & 13.3 & \cite{VGCNBERT2020}\\\cline{2-10}
     & FountaHate & Twitter posts & 4 & 99996 & - & - & - & 15.7 & \cite{VGCNBERT2020} \\\cline{2-10}
     & CoLA & grammar check & 2 & 9594 & 8551 & 1043 & - & 7.7 & \cite{VGCNBERT2020}\\
     \hline
 \end{tabular}
 \end{adjustbox}
 \label{table:datasets}
 \end{center}
 \end{table*}

There are many popular text classification benchmark datasets, while this paper mainly focuses on the datasets used by GNN-based text classification applications. Based on the purpose of applications, we divided the commonly adopted datasets into three types including \textit{Topic Classification}, \textit{Sentiment Analysis} and \textit{Other}. Most of these text classification datasets contain a single target label of each text body. The key information of each dataset is listed in Table \ref{table:datasets}.
\subsubsection{Topic Classification}
\hfill\\
Topic classification models aim to classify input text bodies from diverse sources into predefined categories. News categorization is a typical topic classification task to obtain key information from news and classify them into corresponding topics. The input text bodies normally are paragraphs or whole documents especially for news categorization, while there are still some short text classification datasets from certain domains such as micro-blogs, bibliography, etc. Some typical datasets are listed:
\begin{itemize}
    \item \textit{\textbf{Ohsumed}} \cite{joachims1998text} is acquired from the MEDLINE database and further processed by \cite{TEXTGCN2019} via selecting certain documents (abstracts) and filtering out the documents belonging to multiple categories. Those documents are classified into 23 cardiovascular diseases. The statistics of \cite{TEXTGCN2019} processed Ohsumed dataset is represented in Table~\ref{table:datasets}, which is directly employed by other related works.
    \item \textit{\textbf{R8 / R52}} are two subsets of the Reuters 21587 dataset \footnote{ For the original Reuters 21587 dataset, please refer to this link \url{http://www.daviddlewis.com/resources/testcollections/reuters21578}} which contain 8 and 52 news topics from Reuters financial news services, respectively.
    \item \textit{\textbf{20NG}} is another widely used news categorization dataset that contains 20 newsgroups. It was originally collected by \cite{Lang95}, but the procedures are not explicitly described.
    \item \textit{\textbf{AG News}} \cite{zhang2015character} is a large-scale news categorization dataset compared with other commonly used datasets which are constructed by selecting the top-4 largest categories from the AG corpus. Each news topic contains 30,000 samples for training and 1900 samples for testing.
    \item \textit{\textbf{Database systems and Logic Programming (DBLP)}} is a topic classification dataset to classify the computer science paper titles into six various topics \cite{GTNT2021}. Different from paragraph or document based topic classification dataset, DBLP aims to categorise scientific paper titles into corresponding categories, the average input sentence length is much lower than others.
    \item \textit{\textbf{Dbpedia}} \cite{lehmann2015dbpedia} is a large-scale multilingual knowledge base that contains 14 non-overlapping categories. Each category contains 40000 samples for training and 5000 samples for testing. 
    \item \textit{\textbf{WebKB}} \cite{craven1998learning} is a long corpus web page topic classification dataset.
    \item \textit{\textbf{TREC}} \cite{li2002learning} is a question topic classification dataset to categorise one question sentence into 6 question categories.
\end{itemize}

\subsubsection{Sentiment Analysis}
\hfill\\
The purpose of sentiment analysis is to analyse and mine the opinion of the textual content which could be treated as a binary or multi-class classification problem. The sources of existing sentiment analysis tasks come from movie reviews, product reviews or user comments, social media posts, etc. Most sentiment analysis datasets target to predict the people's opinions of one or two input sentences of which the average length of each input text body is around 25 tokens.
\begin{itemize}
    \item \textit{\textbf{Movie Review (MR)}} \cite{pang2005seeing} is a binary sentiment classification dataset for movie review which contains positive and negative data equally distributed. Each review only contains one sentence.
    \item \textit{\textbf{Stanford Sentiment Treebank (SST)}} \cite{socher2013recursive} is an upgraded version of MR which contains two subsets SST-1 and SST-2. SST-1 provides five fine-grained labels while SST-2 is a binary sentiment classification dataset. 
    \item \textit{\textbf{Internet Movie DataBase (IMDB)}} \cite{maas2011learning} is also an equally distributed binary classification dataset for sentiment analysis. Different from other short text classification dataset, the average number of words of each review is around 221. 
    \item \textit{\textbf{Yelp 2014}} \cite{tang2015document} is a large scale binary category based sentiment analysis dataset for longer user reviews collected from Yelp.com. 
\end{itemize}
Certain binary sentiment classification benchmark datasets are also used by GNN-based text classifiers. Most of them are gathered from shorter user reviews or comments (normally one or two sentences) from different websites including Amazon Alexa Reviews (\textit{\textbf{AAR}}), Twitter US Airline (\textit{\textbf{TUA}}), Youtube comments (\textit{\textbf{SenTube-A}} and \textit{\textbf{SenTube-T}}) \cite{uryupina2014sentube}. 
\subsubsection{Other Datasets}
\hfill\\
There are some datasets targeting other tasks including hate detection, grammaticality checking, etc. For example, \textit{\textbf{ArangoHate}} \cite{arango2019hate} is a hate detection dataset, a sub-task of intend detection, which contains 2920 hateful documents and 4086 normal documents by resampling the merged datasets from \cite{davidson2017automated} and \cite{waseem2016you}. In addition, \cite{founta2018large} proposes another large scale hate language detection dataset, namely \textit{\textbf{FountaHate}} to classify the tweets into four categories including 53,851, 14,030, 27,150, and 4,965 samples of normal, spam, hateful and abusive, respectively. Since there is no officially provided training and testing splitting radio for above datasets, the numbers represented in Table~\ref{table:datasets} are following the ratios (train/development/test is 85:5:10) defined by \cite{VGCNBERT2020}. 

\subsubsection{Dataset Summary}
\hfill\\
Since an obvious limitation of corpus-level GNN models has high memory consumption limitation \cite{TGTRANSFORMER2020,TEXTLEVELGNN2019,HYPERGAT2020}, the datasets with a smaller number of documents and vocabulary sizes such as Ohsumed, R8/R52, 20NG or MR are widely used to ensure feasibly build and evaluate corpus-level graphs. For the document-level GNN based models, some larger size datasets like AG-News can be adopted without considering the memory consumption problem. From Table~\ref{table:datasets}, we could find most of the related works mainly focus on the GNN applied in topic classification and sentiment analysis which means the role of GNNs in other text classification tasks such as spam detection, intent detection, abstractive question answering need to be further exploited. Another observed trend is short text classification are gained less attention compared with long document classification tasks. In this case, GNN in short text classification may be an . 

\subsection{Evaluation Methods}
\subsubsection{Performance Metrics} 
\hfill\\
In terms of evaluating and comparing the performance of proposed models with other baselines, accuracy and F1 are most commonly used metrics to conduct overall performance analysis, ablation studies and breakdown analysis. We use $TP$, $FP$, $TN$ and $FN$ to represent the number of true positive, false positive, true negative and false negative samples. $N$ is the total number of samples. 
\begin{itemize}
    \item \textit{\textbf{Accuracy}} and \textit{\textbf{Error Rate}}: are basic evaluation metrics adopted by many GNN-based text classifiers such as \cite{TEXTGTL2021, liu2016recurrent, DHTG2020, TEXTGCN2019, TGTRANSFORMER2020}. Most of the related papers run all baselines and their models 10 times or 5 times to show the $mean \pm standard$ $deviation$ of accuracy for reporting more convincing results. It can be defined as:
\begin{equation}
  Accuracy = \frac{(TF+TN)}{N},
\end{equation}

\begin{equation}
  ErrorRate = 1-Accuracy = \frac{(FP+FN)}{N}.
\end{equation}

    \item \textit{\textbf{Precision}}, \textit{\textbf{Recall}} and \textit{\textbf{F1}}: are metrics for measuring the performance especially for imbalanced datasets. Precision is used to measure the results relevancy, while recall is utilised to measure how many truly relevant results acquired. Through calculating the harmonic average of Precision and Recall could get F1. Those three measurements can be defined as:
\begin{equation}
  Precision = \frac{TP}{(TP+FP)},
\end{equation}

\begin{equation}
  Recall = \frac{TP}{(TP+FN)},
\end{equation}

\begin{equation}
  F1 = \frac{2 \times Precision \times Recall}{(Precision + Recall)},
\end{equation}

\end{itemize}
Few papers only utilise recall or precision to evaluate the performance \cite{GTNT2021}. However, precision and recall are more commonly used together with F1 or Accuracy to evaluate and analyse the performance from different perspectives  e.g.  \cite{REGNN2019, HGAT2019,VGCNBERT2020,TVGAE2021}. In addition, based on different application scenarios, different F1 averaging methods are adopted by those papers to measure overall F1 score of multi-class (Number of Classes is $C$) classification tasks including:
\begin{itemize}
    \item \textit{\textbf{Macro-F1}} applies the same weights to all categories to get overall $F1_{macro}$ by taking the arithmetic mean. 
    \begin{equation}
     F1_{macro}=   \frac{1}{C}\Sigma_{i=1}^{C} F1_{i},
    \end{equation}
    \item \textit{\textbf{Micro-F1}} is calculated by considering the overall $P_{micro}$ and $R_{micro}$. It can be defined as:
    \begin{equation}
     F1_{micro}=\frac{2 \times P_{micro} \times R_{micro}}{(P_{micro} + R_{micro})}
    \end{equation}
    where:
    \begin{equation}
    P_{micro} = \frac{\Sigma_{i \in C} TP_i}{\Sigma_{i \in C} TP_i+FP_i},
    R_{micro} = \frac{\Sigma_{i \in C} TP_i}{\Sigma_{i \in C} TP_i+FN_i},
    \end{equation}
    \item \textit{\textbf{Weighted-F1}} is the weighted mean of F1 of each category where the weight $W_i$ is related to the number of occurrences of the corresponding $i$th class, which can be defined as:
    \begin{equation}
     F1_{macro}=   \Sigma_{i=1}^{C} F1_{i} \times W_i,
    \end{equation}
\end{itemize}

\subsubsection{Other Evaluation Aspects}
\hfill\\
Since two limitations of GNN-based models are time and memory consumption, therefore, except the commonly used qualitative performance comparison, representing and comparing the GPU or CPU memory consumption and the training time efficiency of proposed models are also adopted by many related studies to demonstrate the practicality in real-world applications. In addition, based on the novelties of various models, specific evaluation methods are conducted to demonstrate the proposed contributions. 
\begin{itemize}
    \item \textit{\textbf{Memory Consumption}}: \cite{HYPERGAT2020, TEXTLEVELGNN2019, DADGNN2021} list the memory consumption of different models for comprehensively evaluating the proposed models in computational efficiency aspect.
    \item \textit{\textbf{Time Measurement}}: \cite{HETEGCN2021, simplegcn} perform performance training time comparison between their proposed models and baselines on different benchmarks. Due to the doubts about the efficiency of applying GNNs for text classification, it is an effective way to demonstrate they could well balance performance and time efficiency.
    \item \textit{\textbf{Parameter Sensitivity}} is commonly conducted by GNNs studies to investigate the effect of different hyperparameters e.g. varying sliding window sizes, embedding dimensions of proposed models to represent the model sensitivity via line chart such as \cite{HGAT2019, HYPERGAT2020, DADGNN2021}. 
    \item \textit{\textbf{Number of Labelled Documents}} is a widely adopted evaluation method by GNN-based text classification models \cite{TEXTGTL2021, DHTG2020, HGAT2019, GTNT2021, TEXTGCN2019, HETEGCN2021, HYPERGAT2020} which mainly analyse the performance trend by using different proportions of training data to test whether the proposed model can work well under the limited labelled training data.
    \item \textit{\textbf{Vocabulary Size}} is similar to the number of labelled documents but it investigates the effects of using different sizes of vocabulary during the GNN training stage adopted by \cite{DHTG2020}.
\end{itemize}

\subsubsection{Metrics Summary}
\hfill\\
For general text classification tasks, Accuracy, Precision, Recall and varying F1 are commonly used evaluation metrics for comparing with other baselines. However, for GNN based models, only representing the model performance cannot effectively represent the multi-aspects of proposed models. In this case, there are many papers conducting external processes to evaluate and analyse the GNN based classifier from multiple views including time and memory consumption, model sensitivity and dataset quantity.

\section{Performance}\label{sec:performance}

\begin{table}[]
\caption{Performance Table. - indicates unavailability. * refers to replication from HyperGAT~\cite{HYPERGAT2020}. }
\small
\begin{tabular}{c|lllllll}
\hline
 \textbf{Type} & \textbf{Method} & \makecell[c]{\textbf{External}\\ \textbf{Resource}} & \textbf{20NG} & \textbf{R8} & \textbf{R52} & \textbf{Ohsumed} & \textbf{MR} \\ \hline
\multirow{12}{*}{\textbf{Corpus-level}} & TextGCN~\cite{TEXTGCN2019} & N/A & 86.34 ± 0.09 & 97.07 ± 00.10 & 93.56 ± 0.18 & 68.36 ± 0.56 & 76.74 ± 0.20 \\ \cline{2-8} 
 & SGC~\cite{SGC2019} & N/A & 88.5 ± 0.1 & 97.2 ± 0.1 & 94.0 ± 0.2 & 68.5 ± 0.3 & 75.9 ± 0.3 \\ \cline{2-8} 
 & S2GC~\cite{S2GCN2020} & N/A & 88.6± 0.1 & 97.4 ± 0.1 & 94.5 ± 0.2 & 68.5 ± 0.1 & 76.7 ± 0.0 \\ \cline{2-8} 
 & TG-transformer~\cite{TGTRANSFORMER2020} & GloVe & - & 98.1±0.1 & 95.2±0.2 & 70.4±0.4 & - \\ \cline{2-8} 
 & DHTG~\cite{DHTG2020} & N/A & 87.13 ± 0.07 & 97.33 ± 0.06 & 93.93 ± 0.10 & 68.80 ± 0.33 & 77.21 ± 0.11 \\ \cline{2-8} 
 & TensorGCN~\cite{TENSORGCN2020} & \makecell[l]{GloVe,\\CoreNLP} & 87.74 ± 0.05 & 98.04 ± 0.08 & 95.05 ± 0.11 & 70.11 ± 0.24 & 77.91 ± 0.07 \\ \cline{2-8} 
 & STGCN~\cite{STGCN2020} & BERT & - & 98.5 & - & - & 82.5 \\ \cline{2-8} 
 & NMGC~\cite{NMGC2021} & N/A & 86.61 ± 0.06 & 97.31 ± 0.09 & 94.35 ± 0.06 & 69.21 ± 0.17 & 76.21 ± 0.25 \\ \cline{2-8} 
 & BertGCN~\cite{BERTGCN2021} & BERT & 89.3 & 98.1 & 96.6 & 72.8 & 86 \\ \cline{2-8} 
 & RobertaGCN~\cite{BERTGCN2021} & RoBERTa & 89.5 & 98.2 & 96.1 & 72.8 & 89.7 \\ \cline{2-8} 
 & T-VGAE~\cite{TVGAE2021} & N/A & 88.08 ± 0.06 & 97.68 ± 0.14 & 95.05 ± 0.10 & 70.02 ± 0.14 & 78.03 ± 0.11 \\ \hline
\multirow{5}{*}{\textbf{Doc-level}} & ReGNN~\cite{REGNN2019} & GloVe & - & 97.93 ± 0.31 & 95.17 ± 0.17 & 67.93 ± 0.33 & 78.71 ± 0.56 \\ \cline{2-8} 
 & Text-Level-GNN~\cite{TEXTLEVELGNN2019} & GloVe & 84.16 ± 0.25* & 97.8 ± 0.2 & 94.6 ± 0.3 & 69.4 ± 0.6 & 75.47 ± 0.06* \\ \cline{2-8} 
 & TextING~\cite{TEXTING2020} & GloVe & - & 98.13 ± 0.12 & 95.68 ± 0.35 & 70.84 ± 0.52 & 80.19 ± 0.31 \\ \cline{2-8} 
 & HyperGAT~\cite{HYPERGAT2020} & N/A & 86.62 ± 0.16 & 97.97 ± 0.23 & 94.98 ± 0.27 & 69.90 ± 0.34 & 78.32 ± 0.27 \\ \cline{2-8} 
 & TextSSL~\cite{TEXTSSL2021} & GloVe & 85.26 ± 0.28 & 97.81 ± 0.14 & 95.48 ± 0.26 & 70.59 ± 0.38 & 79.74 ± 0.19 \\ \hline
\end{tabular}
\label{tab:performance}
\end{table}

While different GNN text classification models may be evaluated on different datasets, there are some datasets that are commonly used across many of these models, including \textbf{20NG}, \textbf{R8}, \textbf{R52}, \textbf{Ohsumed} and \textbf{MR}. The accuracy of various models assessed on these five datasets is presented in Table \ref{tab:performance}. Some of the results are reported with ten times average accuracy and standard derivation while some only report the average accuracy. Several conclusions can be drawn:
\begin{itemize}
    \item Models that use external resources usually achieve better performance than those that do not, especially models with BERT and RoBERTa\cite{BERTGCN2021, STGCN2020}.
    \item Under the same setting, such as using GloVe as the external resource, Corpus-level GNN models (e.g. TG-Transformer\cite{TGTRANSFORMER2020}, TensorGCN\cite{TENSORGCN2020}) typically outperform Document-level GNN models (e.g. TextING\cite{TEXTING2020}, TextSSL\cite{TEXTSSL2021}). This is because Corpus-level GNN models can work in a transductive way and make use of the test input, whereas Document-level GNN models can only use the training data.
    \item The advantage of Corpus-level GNN models over Document-level GNN models only applies to topic classification datasets and not to sentiment analysis datasets such as \textbf{MR}. This is because sentiment analysis involves analyzing the order of words in a text, which is something that most Corpus-level GNN models cannot do.
\end{itemize}

    



\section{Challenges and Future Work}\label{sec:challenges}
\subsection{Model Performance}
With the development of pre-trained models\cite{BERT,liu2019roberta}, and prompt learning methods\cite{gao2021making,liu2021gpt} achieve great performance on text classification. Applying GNNs in text classification without this pre-training style will not be able to achieve such good performance. For both corpus-level and document-level GNN text classification models, researching how to combine GNN models with these pretrained models to improve the pretrained model performance can be the future work. Meanwhile, more advanced graph models can be explored, e.g. more heterogeneous graph models on word and document graphs to improve the model performance.
\subsection{Graph Construction}
Most GNN text classification methods use a single, static-value edge to construct graphs based on document statistics. This approach applies to both corpus-level GNN and document-level GNN. However, to better explore the complex relationship between words and documents, more dynamic hyperedges can be utilized. Dynamic edges in GNNs can be learned from various sources, such as the graph structure, document semantic information, or other models. And hyperedges can be built for a more expressive representation of the complex relationships between nodes in the graph. 
\subsection{Application}
While corpus-level GNN text classification models have demonstrated good performance without using external resources, these models are mostly transductive. To apply them in real-world settings, an inductive learning approach should be explored. Although some inductive corpus-level GNNs have been introduced, the large amount of space required to construct the graph and the inconvenience of incremental training still present barriers to deployment. Improving the scalability of online training and testing for inductive corpus-level GNNs represents a promising area for future work.

\section{Conclusion}\label{sec:conclusion}
This survey article introduces how Graph Neural Networks have been applied to text classification in two different ways: corpus-level GNN and document-level GNN, with a detailed structural figure. Details of these models have been introduced and discussed, along with the datasets commonly used by these methods. Compared with traditional machine learning and sequential deep learning models, graph neural networks can explore the relationship between words and documents in the global structure (corpus-level GNN) or the local document (document-level GNN), giving a good performance. A detailed performance comparison is applied to investigate the influence of external resources, model learning methods, and types of different datasets. Furthermore, we propose the challenges for GNN text classification models and potential future work.
\bibliographystyle{ACM-Reference-Format}
\bibliography{sample-base}

\appendix

\end{document}